\documentclass[10pt,journal]{IEEEtran}
\usepackage{cite}
\usepackage{graphicx}
\usepackage{amsmath}
\usepackage{amssymb}
\usepackage{algorithmic}
\usepackage{array}
\usepackage{pifont}
\usepackage{multirow}
\usepackage{bm}
\usepackage{float}
\usepackage{subfig}
\usepackage{booktabs}
\usepackage[pagebackref=false,breaklinks=true,colorlinks, bookmarks=false]{hyperref}
\usepackage{colortbl}

\graphicspath{{./Imgs/}{./Imgs/Statistics/}}
\newcommand{\figref}[1]{Fig.~\ref{#1}}

\newcommand{\revise}[1]{{\textcolor{black}{#1}}}

\def\ie{\emph{i.e.,~}}
\def\eg{\emph{e.g.,~}}

\def\etc{\emph{etc}}
\def\etal{{\em et al.}}

\usepackage{silence}
\begin{document}
%
\title{Salient Object Detection in Traffic Scene through the TSOD10K Dataset}
\author{Yu Qiu\IEEEauthorrefmark{1},
        Yuhang Sun\IEEEauthorrefmark{1},
        Jie Mei,
        Lin Xiao,
        and Jing Xu
\thanks{Yiu Qiu and Lin Xiao are with the College of Information Science and Engineering, Hunan Normal University, Changsha, 410081, China.}
\thanks{Yuhang Sun and Jing Xu are with the National Key Laboratory of Intelligent Tracking and Forecasting for Infectious Diseases, College of Artificial Intelligence, Nankai University, Tianjin, 300350, China}
\thanks{Jie Mei is with the National Engineering Research Center of Robot Visual Perception and Control Technology, School of Robotics, Hunan University, Changsha, 410082, China.}
\thanks{*Both authors contributed equally to this research.}
\thanks{Jie Mei is the corresponding author (E-mail: jiemei@hnu.edu.cn).}
}

\markboth{Journal of \LaTeX\ Class Files,~Vol.~, No.~, ~2025}%
{Qiu \MakeLowercase{\textit{et al.}}: 
	Salient Object Detection in Traffic Scene through the TSOD10K Dataset}

\maketitle

\begin{abstract}
Traffic Salient Object Detection (TSOD) aims to segment the objects critical to driving safety by combining semantic (\eg collision risks) and visual saliency. Unlike SOD in natural scene images (NSI-SOD), which prioritizes visually distinctive regions, TSOD emphasizes the objects that demand immediate driver attention due to their semantic impact, even with low visual contrast. This dual criterion, \textit{i.e.}, bridging perception and contextual risk, re-defines saliency for autonomous and assisted driving systems.
To address the lack of task-specific benchmarks, we collect the first large-scale TSOD dataset with pixel-wise saliency annotations, named \textbf{TSOD10K}. TSOD10K covers the diverse object categories in various real-world traffic scenes under various challenging weather/illumination variations (\eg fog, snowstorms, low-contrast, and low-light).
Methodologically, we propose a Mamba-based TSOD model, termed \textbf{Tramba}. Considering the challenge of distinguishing inconspicuous visual information from complex traffic backgrounds, Tramba introduces a novel Dual-Frequency Visual State Space module equipped with shifted window partitioning and dilated scanning to enhance the perception of fine details and global structure by hierarchically decomposing high/low-frequency components. To emphasize critical regions in traffic scenes, we propose a traffic-oriented Helix 2D-Selective-Scan (Helix-SS2D) mechanism that injects driving attention priors while effectively capturing global multi-direction spatial dependencies. 
We establish a comprehensive benchmark by evaluating Tramba and 22 existing NSI-SOD models on TSOD10K, demonstrating Tramba's superiority.
Our research establishes the first foundation for safety-aware saliency analysis in intelligent transportation systems.

\end{abstract}

\begin{IEEEkeywords}
Salient object detection, traffic salient object detection, Mamba, visual state space
\end{IEEEkeywords}

\IEEEpeerreviewmaketitle

\section{Introduction}
\label{sec:intro}
\IEEEPARstart{T}he increasing number of vehicles on the road has led to complex and congested traffic environments \cite{luo2018mio,jiang2024traffic}. Accurately identifying and segmenting the salient objects (\eg pedestrians, vehicles, and obstacles) with the potential for traffic accidents plays a pivotal role in quickly taking the measures such as avoidance, deceleration, or braking. Despite the human visual system can automatically focus on important objects in complex and dynamic scenes, it often struggles to react promptly to unexpected and sudden events \cite{jia2023tfgnet}. Therefore, using computer vision techniques to parse the traffic images and outline the area of salient objects can effectively support the ADAS in making adaptive risk avoidance responses.

\begin{figure}[!tb]
    \centering
    \includegraphics[width=\linewidth]{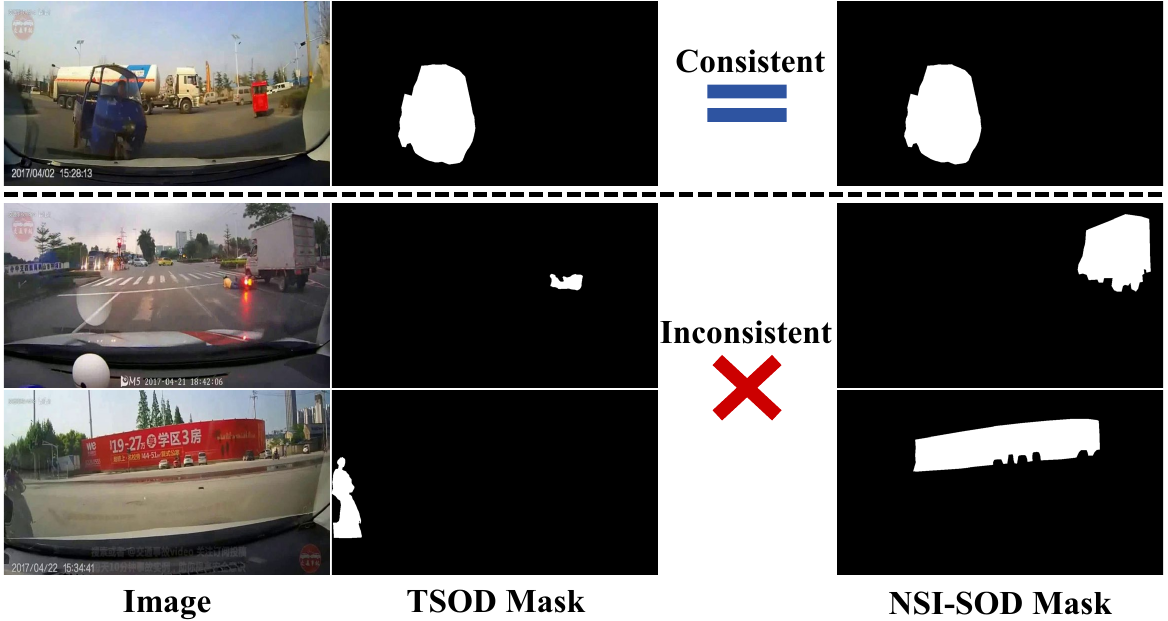}
    \caption{Task discrepancy visualization. Row 2\&3 shows that the salient objects in traffic scenes are sometimes not visually significant, indicating that the TSOD task is driven by both semantic and visual factors.}
    \label{fig:withSOD}
\end{figure}

\begin{figure*}[!tb]
    \centering
    \includegraphics[width=\linewidth]{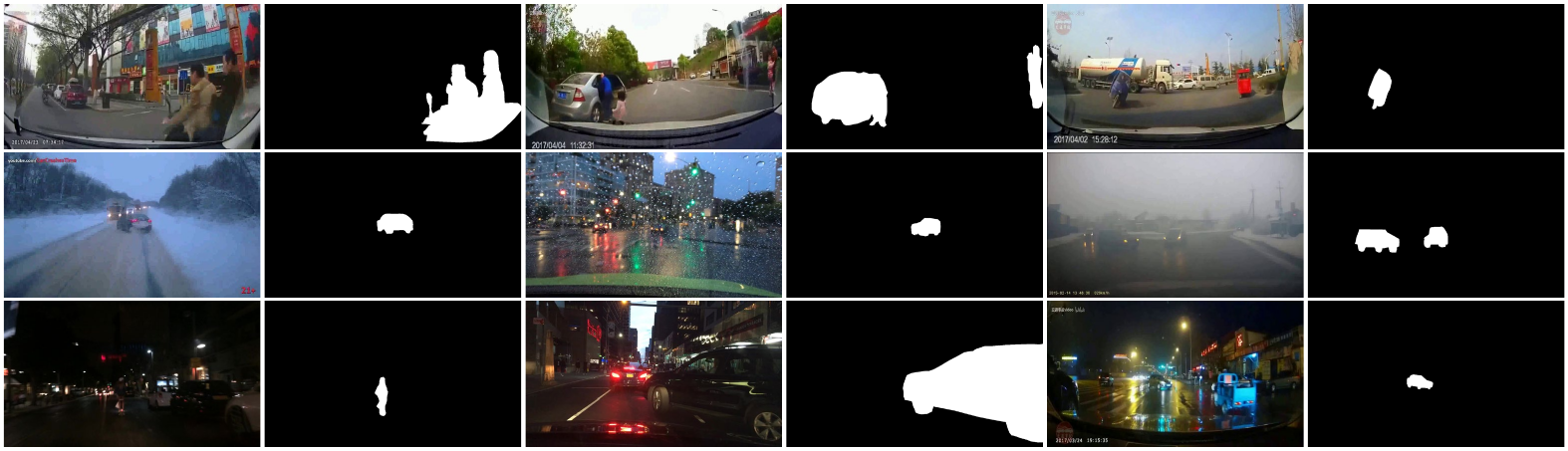}
    \caption{Some traffic images and their corresponding salient object labels picked from the TSOD10K dataset, covering diverse situations: non-motorized vehicles suddenly appearing, motion blur of pedestrians, inconspicuous vehicles suddenly about to overturn, poor visibility in extreme weather conditions/low-light, and glare caused by direct sunlight or car headlights.}
    \label{fig:examples}
\end{figure*}

Salient Object Detection (SOD) focuses on segmenting perceptually dominant objects in natural scene images (NSI) through photometric priors (\eg color contrast, edge density) \cite{qiu2024superpixel, SOD_ICON, vst}. While some studies \cite{jia2023tfgnet} have extended this paradigm to traffic scenes via visual saliency modeling, the fundamental limitations persist: 1) Traditional visual saliency methods disproportionately emphasize photometrically distinctive regions while overlooking safety-critical yet visually ambiguous targets (\eg fog-occluded pedestrians), thus undermining risk-aware perception in autonomous driving; 2) Previous attempts \cite{jia2023tfgnet} are limited due to relying on a sub-dataset composed of traffic scene images filtered from the general SOD datasets, leading to incomplete coverage of safety-relevant edge cases (\eg abrupt jaywalking, adverse weather hazards). This gap in task definition and benchmarking necessitates a dedicated framework for traffic-oriented saliency analysis.
Besides, driver attention prediction (DAP) methods attempt to model saccadic gaze patterns to prevent hazardous operations \cite{DAP2}. However, such gaze-driven paradigms prioritize transient fixation points over holistic object-level awareness.
Moreover, TSOD reduces the reliance on drivers to notice and react to emergencies, thereby significantly decreasing the chances of accidents caused by human error, such as inattention or incorrect hazard assessment.

The absence of open benchmarks for traffic saliency analysis has severely constrained methodological exploration in this safety-critical domain. To bridge this gap, we present the first large-scale dataset specifically designed for traffic salient object detection, named TSOD10K, comprising 13,753 vehicle-captured images with pixel-precise annotations. 
TSOD10K spans various real-world traffic scenes, including urban intersections, highways, rural roads, and parking lots, captured under distinct weather/illumination combinations (\eg rain, snow, fog, sunny, and low-light). 
To ensure the quality of labeling, we ensure a three-stage workflow involving annotation by safety-aware driver (5+ years of driving experience), cross-validation by traffic engineers, and edge-case arbitration via simulated risk assessment. For the convenience of use, we set three emergency levels (\ie Normalcy, Caution, Crisis) based on risk parameters including the object distance, angle, behavior, visual visibility, and environment. As illustrated in \figref{fig:examples}, TSOD10K uniquely encodes both visual saliency and implicit risk semantics. For instance, pedestrians who are almost invisible in low-light environments but pose a high risk to traffic safety are labeled as high priority. Overall, the characteristics of our TSOD10K dataset are summarized as follows: \textbf{1)} the first large-scale annotated TSOD dataset; \textbf{2)} high diverse traffic scenes; \textbf{3)} containing the majority of common targets on roads; \textbf{4)} dynamic risk stratification of downstream services is supported.

The core challenge of TSOD lies in the semantic-visual discrepancy of traffic scenes: safety-critical objects often exhibit minimal photometric contrast against complex backgrounds (\eg fog-occluded pedestrians), rendering conventional appearance-based methods ineffective. Recently, a novel state space model (SSM), Mamba \cite{mamba}, has gained popularity for its ability to model global context like Transformer while maintaining dynamic weights with linear complexity. Moreover, although natural images encode rich frequency semantics (high frequencies capturing fine-grained textures and low frequencies preserving structural continuity) \cite{sun2025frequency}, existing SSMs struggle to model cross-scale frequency interactions due to their inherent limitations in maintaining spatial locality and multi-resolution relationships. We introduce \textbf{Tramba}, a novel Mamba-based architecture for TSOD that redefines traffic saliency analysis through frequency-aware visual state spaces. 
Tramba overcomes these constraints through its Dual-Frequency Visual State Space (DFVSS) module, which synergistically integrates RGB and frequency domains via two complementary mechanisms: 1) Window-SS2D which is inspired by local window attention can employ non-overlapping windows to preserve high-frequency details, and 2) Dilation-SS2D adapts dilated convolution principles to establish long-range dependencies to recover low-frequency structural contexts. 
On the other hand, Tramba incorporates a Helix-SS2D mechanism in its decoder path, specifically engineered to model directional saliency in driving environments. Unlike conventional unidirectional scanning in SSMs, Helix-SS2D implements a multi-perspective feature aggregation strategy that simulates driver attention priors of an innate tendency to prioritize specific spatial orientations through a helical scanning pattern.

Our main contributions are as follows:

\begin{itemize}
    \item We propose \textbf{TSOD10K}, the first large-scale traffic saliency dataset with pixel-level safety-semantic annotations, which encompasses 13,753 challenging real-world traffic scenarios and common traffic objects and is equipped with various hazard stratification.
    \item We present \textbf{Tramba}, a Mamba-inspired model featuring a Dual-Frequency Visual State Space (DFVSS) module. 
    \item We develop an HVSS module with a traffic-oriented Helix-SS2D mechanism, which integrates driving attention priors to emphasize critical perspectives while enhancing directional understanding of traffic scenes.
    \item We establish the first comprehensive benchmark for advancing future traffic saliency research. Extensive experiments demonstrate the superiority of Tramba in traffic scene parsing.
\end{itemize}

\section{Related Work}
\label{sec:related}
\begin{figure*}[!tb]
\centering
\includegraphics[width=\linewidth]{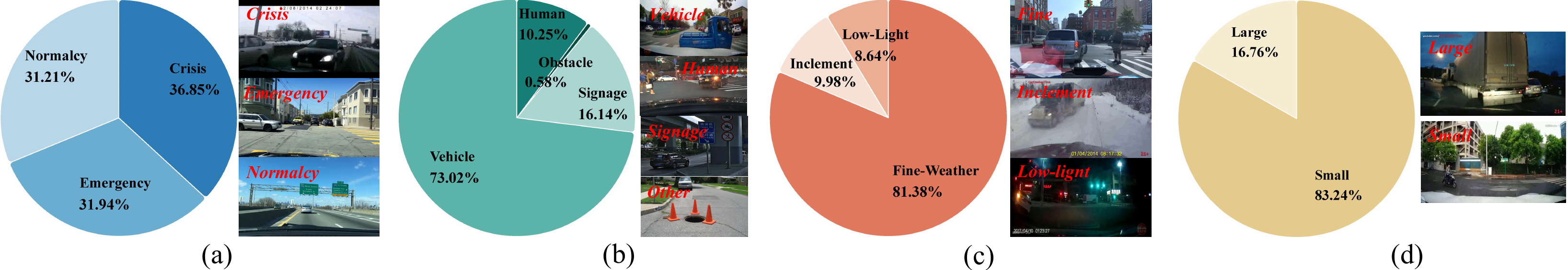}
\caption{Data statistics of our TSOD10K dataset. (a) Risk stratification: proportions of Normalcy, Emergency, and Crisis scenarios; (b) Object category prevalence including Vehicle, Human, Signage, and Obstacles; (c) Environmental condition distribution including Fine-Weather, Inclement, and Low-Light; (d) Target size analysis on Large vs. Small objects.}
\label{fig:statistics}
\end{figure*}

\begin{figure}[!tb]
\centering
\subfloat[]{\includegraphics[width=.38\linewidth] {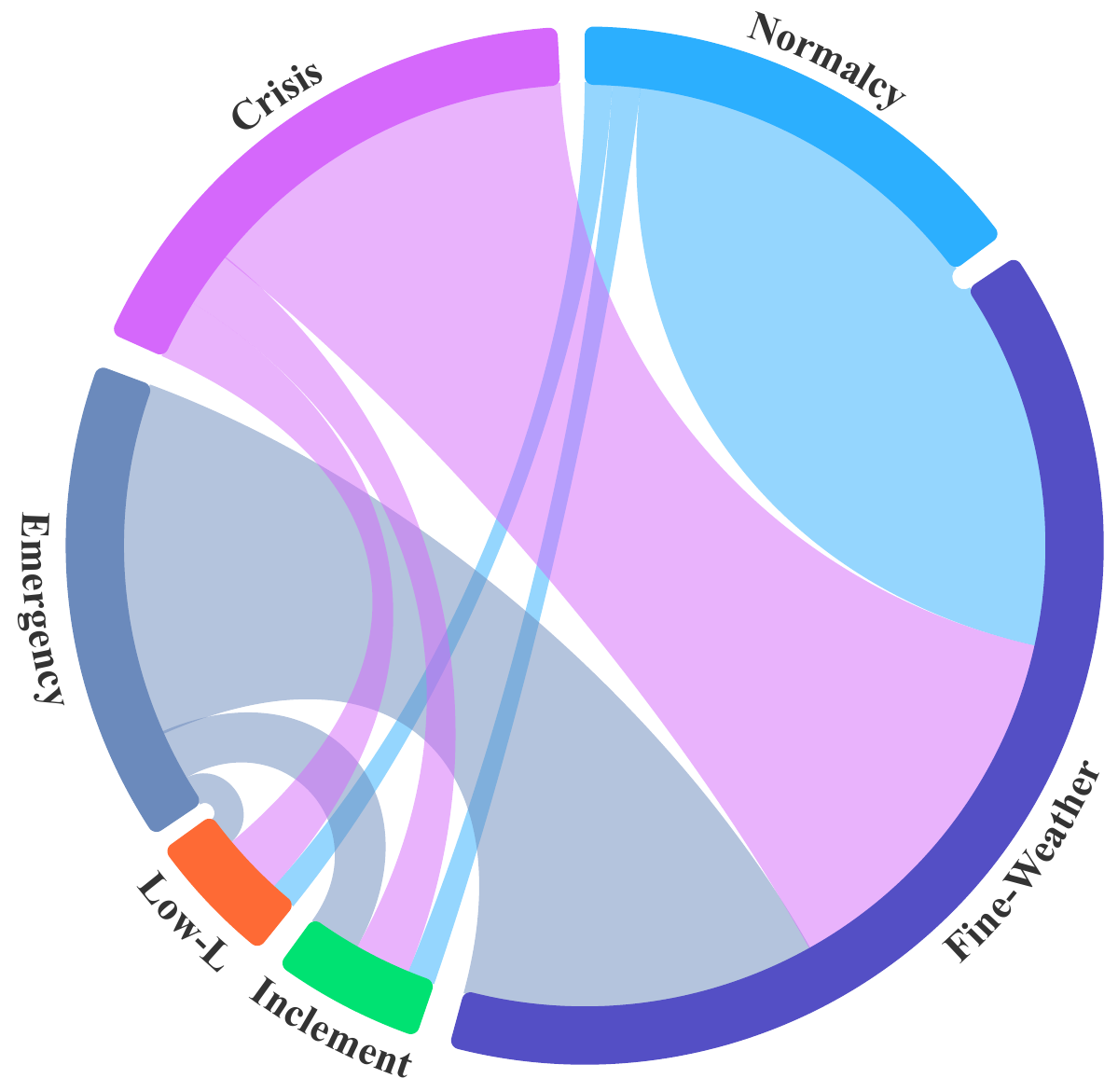}\label{fig:statistic_size}} \hspace{0.2in}
\subfloat[]{\includegraphics[width=.49\linewidth]{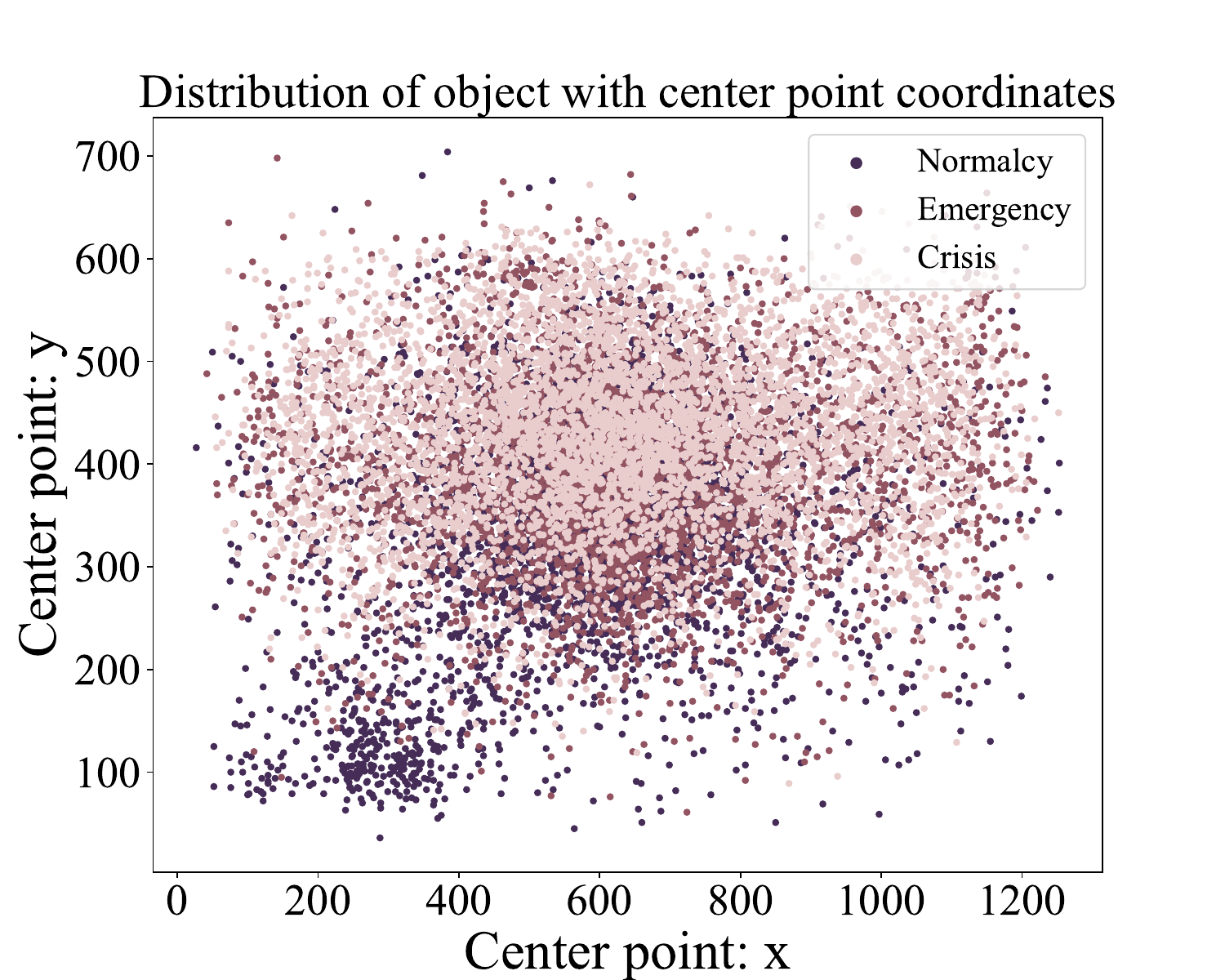}\label{fig:statistic_number}}
\caption{Data statistic of our TSOD10K on attribute dependence and object location. (a): Multi-dependencies among attributes, with larger arc lengths indicating higher correlation probabilities. (b): Locations of traffic salient objects' center points.}
\label{fig:attributes}
\end{figure}

\subsection{Salient Object Detection}
Salient object detection (SOD) aims to imitate the human visual perception system to capture the most significant regions in an image \cite{SOD_VST,SOD_ICON, SOD_TIP2, SOD_TIP3, SOD_TIP4, SOD_TIP5}. From 2015 to 2022, CNN-based networks have remarkably boosted SOD performance due to their powerful multi-level feature learning capability \cite{qiu2022a2sppnet}. 
In recent years, Transformer-based models have gradually become the mainstream framework of SOD \cite{SOD_ICON, xie2024csfwinformer}. VST \cite{SOD_VST} first employs the Transformer-based encoder for SOD.
However, they all endure the complexity of feature processing using transformers, which is quadratic proportional to the length of the image patch.
Moreover, some studies have attempted to expand SOD to other visual scenes, such as multi-modal SOD \cite{RGB-D_SOD}, SOD for remote sensing images \cite{SOD_DHCont}, and traffic SOD \cite{jia2023tfgnet}. Specially, Jia \etal \cite{jia2023tfgnet} proposed the TFGNet for traffic SOD. 
However, as shown in Fig. \ref{fig:withSOD}, the significance of traffic scenes is not completely consistent with that of natural scenes. In the context of a traffic driving environment, the most salient objects are invariably the most critical to road safety, as they possess the capacity to capture and command the driver's attention. Hence, these existing methods based on visual saliency for traffic scenes focus excessively on visually distinctive objects while neglecting safety-critical but low-contrast targets, highlighting the need for a dedicated traffic saliency analysis framework.

\subsection{Driving Attention Prediction}
Driver attention is one of the major factors for roadway safety. Driving Attention Prediction (DAP) aims to design human-centric ADAS to parse the driver's attention-sensing behaviors and predict unsafe maneuvers \cite{DAP2}. 
The current publicly available DAP datasets mainly include BDD-A \cite{BDD-A}, DADA-2000 \cite{fang2021dada}, and DR(eye)VE \cite{drve}. 
Based on these datasets, lots of DAP models have been proposed \cite{fang2021dada}.
For example, Fang \etal \cite{fang2021dada} proposed a new SCAFNet for DAP, where the semantic context feature of the driving scene is introduced to help the finding of the key objects/regions that attract drivers' attention. Gan \etal \cite{gan2022multisource} proposed the adaptive driver attention (ADA) model integrating domain adaptation and attention mechanism to predict salient regions in different traffic scenes.
However, \textbf{1)} due to the differences in driving habits and safety awareness, drivers' fixation points are subjective, and \textbf{2)} DAP task typically outputs the heatmaps of visual attention points rather than the pixel-level predictions of complete objects, limiting their applicability in the complex traffic scenes. Instead, TSOD reduces the reliance on drivers to notice and react to emergencies, thereby significantly decreasing the chances of accidents caused by human inattention.

\subsection{Scanning Mechanism in Visual Mamba}
As a crucial component of Mamba in vision, effective scanning mechanisms can enhance model performance and facilitate the training process \cite{qu2024survey}. Especially, Cross Scan \cite{liu2024vmamba} is the most widely adopted, which flattens patches along four distinct paths and can be regarded as a fusion of two bidirectional scans. Moreover, the continuous scanning mechanism \cite{he2024mambaad} processes adjacent tokens between columns (or rows), instead of traveling to the opposite tokens in Cross Scan. Hilbert Scan \cite{he2024mambaad} travels a sinuous path based on the Hilbert matrix. In addition, efficient scanning mechanism \cite{pei2024efficientvmamba} proceeds images by skipping patches to accelerate the training and inference process. In contrast to the aforementioned flattened scan methods, stereo-scan methods excel in capturing a broader spectrum of knowledge during the scanning process \cite{qu2024survey}. For example, Hierarchical Scan \cite{chen2024mim} uses various kernel sizes to capture the semantic knowledge from global to local or from macro to micro perspectives. Spatiotemporal scan \cite{yao2024spectralmamba} contains two 3D scans, \ie, spatial-first scanning and temporal-first scanning. Numerous researches also aim to combine various scanning methods to achieve comprehensive feature modeling \cite{dong2024fusion}.

\section{TSOD10K Dataset}
\label{sec:dataset}

\subsection{Data Collection and Annotation}\label{ssec:DataCollection}
To facilitate the community research, we build the first dataset specifically for the traffic salient object detection, named TSOD10K.
Overall, the 13,753 images of our TSOD10K dataset are mainly obtained by extracting video frames from 1) a general driving video dataset BDD100K \cite{BDD100K}, 2) three driver attention prediction video datasets DADA-2000 \cite{fang2021dada}, DR(eye)VE \cite{DR(eye)VE} and CDNN \cite{TrafficSalency}, and 3) an accident anticipation video dataset CCD \cite{CCD}.
The detailed descriptions of these five datasets are as follows:

\textbf{DADA-2000} \cite{fang2021dada} dataset is a large-scale driving attention dataset focused on traffic accidents within the driver's field of view. It consists of 2,000 driving accident sequences, totaling 658,476 frames, and covers various weather conditions, lighting conditions, and road environments.

\textbf{DR(eye)VE} \cite{DR(eye)VE} dataset is designed for driving attention tasks and comprises 74 video sequences, each approximately 5 minutes long, totaling over 500,000 frames with fixation point labels.

\textbf{BDD100K} \cite{BDD100K} is the largest and most diverse general traffic video dataset, consisting of 100,000 videos. Each video is approximately 40 seconds long, with a resolution of 720p at 30 fps. The dataset encompasses different weather conditions (\eg sunny, cloudy, and rainy), and different times of the day (\eg daytime and nighttime).

\textbf{CCD} \cite{CCD} is an accident anticipation video dataset focused on traffic accident analysis, including 1,500 cropped videos from YouTube (each video contains 50 frames, captured over 10 seconds) and 3,000 regular videos from BDD100K. CCD features diverse accident annotations, encompassing environmental attributes (day/night, snowy/rainy/clear conditions). 

\textbf{CDNN} \cite{TrafficSalency} is an eye-tracking dataset obtained from 28 experienced drivers watching 16 traffic-driving videos, comprising a total of 70,000 frames. 
For every traffic video, we sample frames at ten-frame intervals and manually filter out similar frames. Specifically, for accident videos, we only extract 1$ \sim $3 frames that include accident targets, covering pre-incident, critical, and post-incident stages of accidents.

To ensure label accuracy and reliability, we use a triple-check annotation strategy with annotators who have extensive driving experience and strong safety awareness. The first annotator labels object regions independently, carefully checks the annotations, and re-annotates the unsatisfactory regions. The second annotator re-checks the annotations and discusses any disagreements with the first annotator for re-annotation. Finally, the third annotator re-checks the annotations. If there are disagreements, these three annotators discuss and make re-annotations again.

\subsection{Data Statistics}\label{ssec:DataStatistics}
For convenience, we classify all images into different categories according to their various attributes:
\begin{itemize} 
\item \textit{Emergency levels.} We select three experienced drivers to manually assign an emergency level (\ie crisis, emergency, and normalcy) to each image by considering factors such as distance, angle, direction, behavior, visual visibility, and environment. \figref{fig:statistics}(a) displays the proportion of images at different emergency levels. 
\item \textit{Object categories.} TSOD10K includes four common traffic object categories (human, vehicle, signage, and other obstacle). \figref{fig:statistics}(b) presents the proportion of each categories. Specially, other obstacles include animals, plants, trash cans, roadblocks, tires, road debris, \etc. The vehicles are the most common category, followed by human and road signage. Specially, this paper classifies the people riding bicycles or electric bikes as vehicles.
\item \textit{Weather conditions.} The weather conditions have a significant impact on traffic situations, so we divide all images into three categories based on weather conditions (fine weather, inclement weather, and low-light) as shown in \figref{fig:statistics}(c). Most images are captured in fine weather.
\item \textit{Object sizes.} Based on the ratio of object size in the entire image, we divide all objects into two categories: large (ratio $\geq$ 0.1) and small (ratio $<$ 0.1), as shown in \figref{fig:statistics} (d). We can see that most objects in TSOD10K are small. 
\end{itemize}

We then display the co-attribute dependencies in \figref{fig:attributes}(a), showing that critical and emergency situations are more likely under inclement weather and low-light conditions, which also aligns with our subjective perspective.
We plot the locations of center points of objects in \figref{fig:attributes}(b). Obviously, 1) objects are substantially randomly distributed, indicating the universality of TSOD10K, and 2) the number of objects near the center of the image is the highest, indicating that the view center of the driver is still the most important.

\subsection{Data Name and Split}\label{ssec:DataSplit}
In order to unify and clarify usage, we rename all images according to their various attributes. The detailed naming rule is ``C/E/N\_H/V/S/O\_F/I/L\_L/S\_\textit{ID}.jpg'', where ``C/E/N'' refers to the emergency levels, ``H/V/S/O'' refers to the object categories, ``F/I/L'' refers to the weather conditions, ``L/S'' refers to the object sizes, and ``\textit{ID}'' is a five number.
We randomly split TSOD10K into a training set and a testing set at an 8:2 ratio, naming them TSOD10K-TR (10,997 image-mask pairs) and TSOD10K-TE (2,756 pairs). To ensure the diversity of objects, the 8:2 split is applied to each sub-class (human, vehicle, signage, and others).

\begin{figure*}[!tb]
    \centering
    \includegraphics[width=\linewidth]{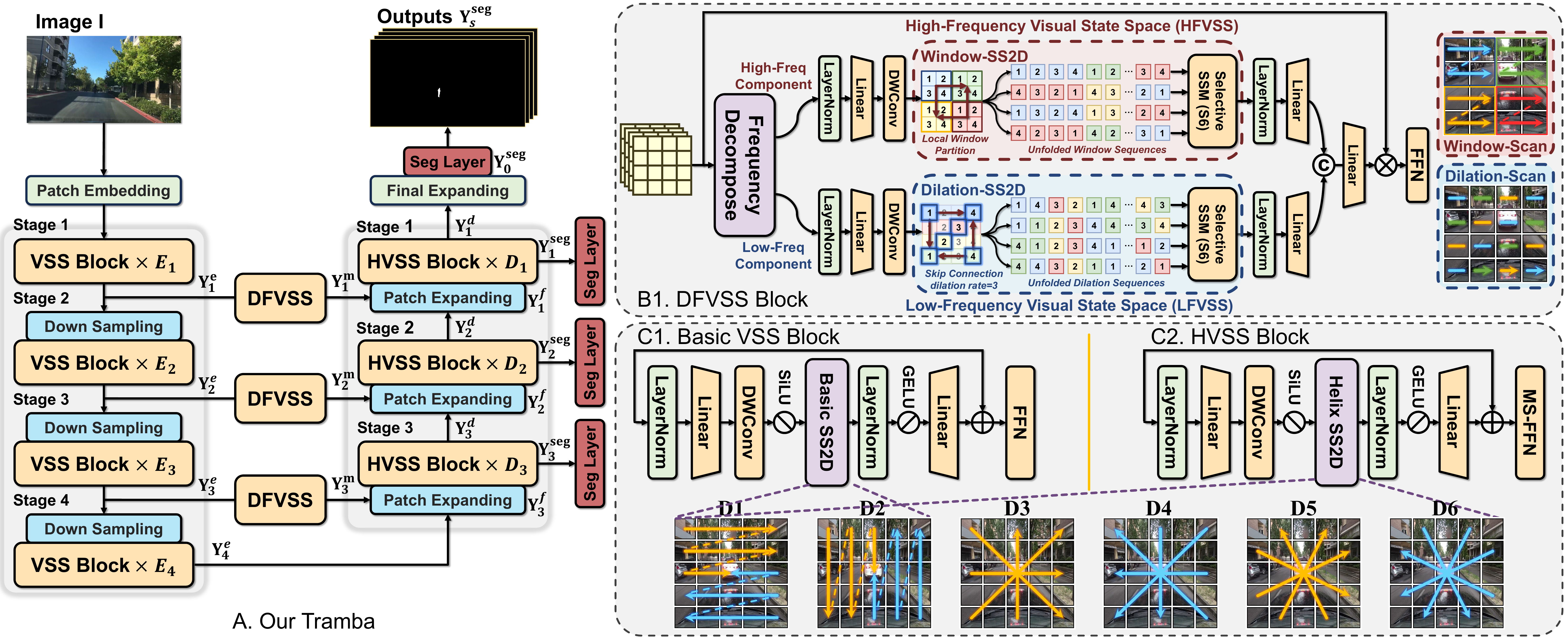}
    \caption{\textbf{Detailed illustration of our Tramba.} \textbf{A:} Tramba adopts a U-shaped encoder-decoder architecture, with the encoder built on VMamba-based Visual State Space (VSS). \textbf{B: } Our Dual-Frequency VSS (DFVSS) \ref{sec:dfvss} decouples encoded features into frequency domains via DCT, utilizing the high-frequency VSS (red component) with a sliding local window mechanism for fine-grained details, and a low-frequency VSS (blue components) with a dilated leapfrog scanning mechanism for global contextual dependencies. \textbf{C1-C2:} The basic VSS blocks incorporate horizontal and vertical scanning (D1-D2), while our Helix-VSS (HVSS) \ref{sec:hvss} introduces a center-focused Helix scanning strategy (D3-D6) for driving-centric perception.
    }
    \label{fig:overal}
\end{figure*}

\section{Method}
\label{sec:method}

\subsection{Preliminaries}

Recently, State Space Models (SSMs) \cite{S4, mamba} have been proposed as an alternative due to linear complexity in sequence modeling, with concepts derived from control engineering. To handle discredited input sequence ${x}=\left ( {x_{0}}, {x_{1}},\dots, {x_{L}} \right )$, Continuous-time SSM is discretized by Zero-Order Hold (ZOH). Discrete-time SSM, \eg Structured SSM (S4) \cite{S4}, can be expressed as follows:
\begin{equation}
\begin{split}
    h_{k} &= \overline{A} h_{k-1} + \overline{B} x_{k}, \\
    y_{k} &= C h_{k},
    \label{e1}
\end{split}
\end{equation}
This recurrent structure can also be reformulated as a 1D convolutional form:
\begin{equation}
\begin{split}
    y  = x \ast \overline{K},\ 
    \overline{K}=\left ( C \overline{B},C \overline{A}\overline{B} ,\dots , C\overline{A}^{L-1}\overline{B}  \right ),
    \label{e2}
\end{split}
\end{equation}
where $\overline{K}$ denotes an input-independent structured convolutional kernel that can be precomputed, enabling parallelization while limiting dynamic inference.
To make SSM input-dependent, Mamba (Selective SSM, S6) \cite{mamba} integrates a selection mechanism, allowing selective propagation or forgetting of information along the sequence. Expanding into vision, VMamba \cite{vmamba} uses a 2D-Selective-Scan (SS2D) mechanism, which cross-scans 2D input $\mathbf{X}\in\mathbb{R}^{H \times W \times C}$ into different sequences for concurrent SSM processing and merges the output sequences back to the original size, as $\overline{\mathbf{X}} \in \mathbb{R}^{H \times W \times C}$. SS2D mechanism can be represented as:
\begin{small}
\begin{equation}
\begin{split}
    \overline{\mathbf{X}} = \mathrm{Merge}(\mathrm{SSM}(\mathrm{Scan}(\mathbf{X}))).
    \label{e3}
\end{split}
\end{equation}
\end{small}

\begin{figure*}[!tb]
    \centering
    \includegraphics[width=\linewidth]{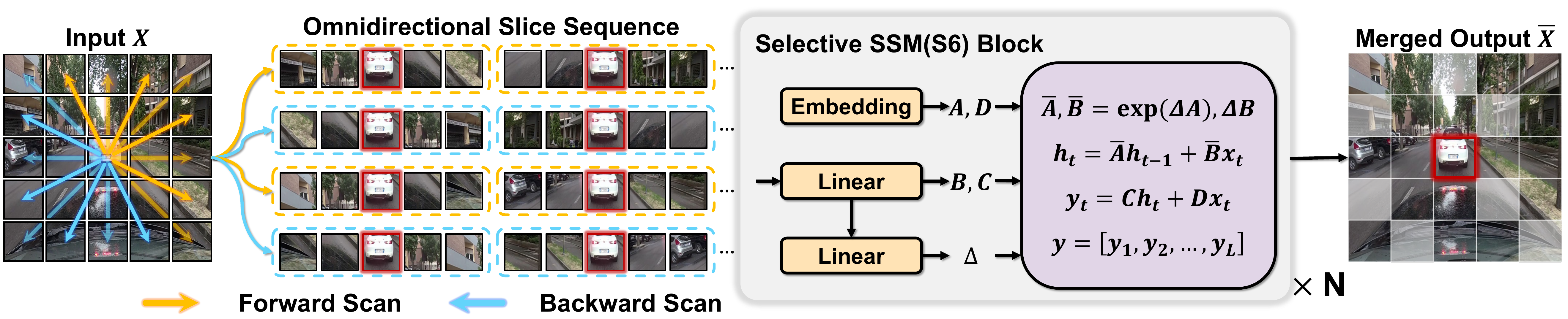}
    \caption{Details of our Helix SS2D. Helix SS2D scans short sequences along a central axis, creating forward-backward slices that cover the central region. S6 blocks capture and merge the contextual information from these sequences into the output $\overline{\mathbf{X}}$.}
    \label{fig:helix}
\end{figure*}

\subsection{Network Architecture}
Fig. \ref{fig:overal} (A) shows the architecture of Tramba. Based on the U-shaped Mamba-based structure, Tramba primarily consists of three components: 1) a four-stage Visual State Space (VSS) encoder based on the VMamba backbone \cite{vmamba} for multi-level feature extraction, 2) three Dual-Frequency Visual State Space Module (DFVSS) in the skip connections that separates and guides the rich high- and low-frequency patterns in the encoder features, and 3) a three-stage Helix VSS (HVSS) that fuses hierarchical features and progressively decodes for refined segmentation maps.

Input image $\mathbf{I}\in\mathbb{R}^{H\times W\times3}$ in the encoder stage is first patched through $4\times 4$ embedding without positional encoding. Each encoder stage $s \in \left \{ 1,2,3,4 \right\}$ comprises a $2 \times$ downsampling layer and $E_{s}$ stacked VSS blocks to extract multi-level features $\mathbf{Y}_{s}^{e}$, with resolution ratios $\frac{1}{4}$, $\frac{1}{8}$, $\frac{1}{16}$, and $\frac{1}{32}$, respectively.
The DFVSSs are cascaded after the first three encoder stages, obtaining intermediate features $\mathbf{Y}_{s}^{m}$ at the original resolution, enhanced by refined frequency-domain information. On the other side, each decoder stage $s$ ($s \in \left \{3,2,1 \right\}$) stacks $D_{s}$ HVSS blocks. The input to each decoder stage is the fused features $\mathbf{Y}_{s}^{f}$ obtained by concatenating the high features $\mathbf{Y}_{s+1}^{d}$ from the skip connections with the upsampled intermediate-level features $\mathbf{Y}_{s}^{m}$.
Each Seg layer (head module) consists of a single $1\times 1$ convolution to map $\mathbf{Y}_{s}^{d}$ to the prediction $\mathbf{Y}_{s}^{seg}$ for deep supervision. The final prediction $\mathbf{Y}_{0}^{seg}$ is obtained through a Final Expanding layer with $4\times$ upsampling.

\subsection{Dual-Frequency Visual State Space}
\label{sec:dfvss}
Due to similarity and low-contrast with background, accurately locating and segmenting emergency objects in complex traffic environments is particularly challenging, especially under adverse weather or low-light scenarios. Natural images encompass rich frequency information that encodes image variations: high frequencies capture fine local details, while low frequencies convey global structures.
\revise{Considering the distinguishability of foreground and background features in the frequency domain, we propose a novel Dual-Frequency Visual State Space (DFVSS) module to combine RGB's and frequency's complementary advantages.}
Specifically, DFVSS first introduces the offline discrete cosine transform (DCT) to map RGB features into the frequency domain, initially decomposing them into coarse-grained high/low-frequency components. To enhance the fine-grained frequency representation, DFVSS contains two parallel branches to model frequency information interaction separately: High-Frequency VSS (HFVSS) with Window-SS2D and Low-Frequency VSS (LFVSS) with Dilation-SS2D.

Specifically, the HFVSS branch with Window-SS2D mechanism introduces a novel Window Scan, which pre-scans non-overlapping shifted windows (\eg $4\times 4$ or $8\times 8$) horizontally or vertically to generate local patch sequences. In this way, the structural relationships within the image patch local sequences are prioritized, which enhances the capability of capturing rapidly changing local edges and texture features.
Unlike local window self-attention\cite{SwinTransformer}, HFVSS facilitates information interaction between local windows without sliding, effectively mitigating structural information loss and receptive field degradation.
For 2D input $\mathbf{X} = \left \{ \mathbf{x}_{ij}\in\mathbb{R}^{C} | i \in [0,H-1], j \in [0,W-1] \right \}$, the process of Window Scan can be formalized as:
\begin{equation}
\begin{split}
    &\overrightarrow{\mathbf{X}_{\mathrm{w}}} = \!\!\!\!\bigcup_{i=0}^{\left\lfloor \frac{H}{S} \right\rfloor - 1}\! \bigcup_{j=0}^{\left\lfloor \frac{W}{S} \right\rfloor - 1}\!\!\!\!  \left\{ \mathbf{X}_{iS + u, jS + v} \mid (u, v) \in [S]^2 \right\},\\
    &\overrightarrow{{\mathbf{X}_{\mathrm{w}}}'} = \!\!\!\! \bigcup_{j=0}^{\left\lfloor \frac{W}{S} \right\rfloor - 1} \bigcup_{i=0}^{\left\lfloor \frac{H}{S} \right\rfloor - 1}\!\! \left\{ \mathbf{X}_{iS + u, jS + v} \mid (v, u) \in [S]^2 \right\},\\
    &\overleftarrow{\mathbf{X}_{\mathrm{w}}}=\left \{ \overrightarrow{\mathbf{X}_{\mathrm{w}}}_{(N-1-i)} \right \}_{i=0}^{N-1}, \\
    &\overleftarrow{{\mathbf{X}_{\mathrm{w}}}'}=\left \{ \overrightarrow{{\mathbf{X}_{\mathrm{w}}}'}_{(N-1-i)} \right \}_{i=0}^{N-1},
    \label{e10}
\end{split}
\end{equation}
where $S$ denotes the window size, $N$ is equivalent to $HW$, ``$\longrightarrow$" and ``$\longleftarrow$" denote forward-backward scanning.

To further enhance the understanding of global context, we propose a Dilated-SS2D mechanism with Dilated Scan in the LFVSS branch, which prioritizes modeling long-range dependencies between patches through sparse connections. Our dilation scan introduces gaps (or dilation) between the sampled points, allowing an exponentially expanded receptive field at a lower cost. The level of dilation is controlled by the dilation rate. For 2D input $\mathbf{X}$, the process of Dilation Scan can be formalized as:
\begin{equation}
\begin{split}
    &\overrightarrow{\mathbf{X}_{\mathrm{d}}} = \bigcup_{i=0}^{R - 1} \left\{ \text{flatten}\left ( \mathbf{X} \right )_{i+nR}\right\}_{n=0}^{\left\lceil \frac{HW-i}{R}  \right\rceil - 1},\\
    &\overleftarrow{\mathbf{X}_{\mathrm{d}}} = \bigcup_{i=0}^{R - 1} \left\{ \text{flatten}\left ( \mathbf{X}^{\mathrm{T}} \right )_{i+nR}\right\}_{n=0}^{\left\lceil \frac{HW-i}{R}  \right\rceil - 1 },\\
    &\overleftarrow{\mathbf{X}_{\mathrm{d}}}=\left \{ \overrightarrow{\mathbf{X}_{\mathrm{d}}}_{(N-1-i)} \right \}_{i=0}^{N-1}, \overleftarrow{{\mathbf{X}_{\mathrm{d}}}'}=\left \{ \overrightarrow{{\mathbf{X}_{\mathrm{d}}}'}_{(N-1-i)} \right \}_{i=0}^{N-1},
    \label{e11}
\end{split}
\end{equation}
where $R$ denotes the dilation rate, $\mathrm{flatten}$  refers to the operation of unfolding into 1D sequence by rows.
Recognizing the significance of various frequency bands for precise object localization and segmentation, we concatenate and fuse the outputs from the dual-branch structure to allow full interaction between different frequency spectra. This comprehensive frequency domain representation is then used to enhance the RGB features.

\subsection{Helix Visual State Space}
\label{sec:hvss} 
Recently, some studies \cite{he2024mambaad,pei2024efficientvmamba} have focused on enhancing Mamba's 2D processing capabilities by incorporating additional scanning strategies in 2D-Selective-Scan (SS2D). However, these approaches still fall short in fully capturing the rich directional information, and may struggle in traffic scenarios with complex, dynamic backgrounds and diverse target characteristics (as illustrated in Fig. \ref{fig:statistics}). 

\setlength{\tabcolsep}{0.25mm}
\begin{table*}[!t]\small
\centering
\caption{Quantitative comparison results on {TSOD10K-TE} (for TSOD) and {DUTS-TE\cite{DUTS}} (for NSI-SOD) dataset. 
The top three results in each column are highlighted in \textcolor{red}{red}, \textcolor{blue}{blue}, and \textcolor{green}{green}. Only the metric $\mathcal{M}$ is such that lower values indicate better performance.
}
\begin{tabular}{l|c|c|c||cccccccc||cccc} \toprule[2pt]
    \multirow{2}{*}{\textbf{Methods}} 
    & \multirow{2}{*}{\textbf{Publish}}
    & \multicolumn{1}{c|}{\textbf{Params}}
    & \multicolumn{1}{c||}{\textbf{FLOPs}}
    & \multicolumn{8}{c||}{\textbf{TSOD10K-TE (2,756 Image-GT Pairs)}} 
    & \multicolumn{4}{c}{\textbf{DUTS-TE\cite{DUTS} (5,019 Pairs)}} 
    \\ \cline{5-16} 
    &
    &(M)&(G)
    & $F_{\beta}^{adp}$
    & $F_{\beta}^{max}$
    & $F_{\beta}^{mean}$
    & $E_{\xi}^{adp}$
    & $E_{\xi}^{max}$
    & $E_{\xi}^{mean}$
    & $S_{m}$
    & $\mathcal{M}$

    & $F_{\beta}^{w}$
    & $E_{\xi}^{max}$
    & $S_{m}$
    & $\mathcal{M}$

    \\ \hline
    \multicolumn{16}{c}{\textbf{CNN-Based}}
    \\ \hline
    PiCANet
    \cite{picanet}
    & CVPR18 
    &47.22&59.78
    & .5875 & .7160 & .6779 & .7924
    & .8721 & .8198 & .8156 & .0187
    & .8011 & .9096 & .8492 & .0455
    \\
    CPD-R 
    \cite{cpd}
    & CVPR19 
    &47.85&17.82
    & .7334 & .8190 & .7971 & .8857
    & .9228 & .8992 & .8829 & .0121
    & .7871 & .9087 & .8576 & .0480
    \\
    EGNet-R 
    \cite{egnet}
    & CVPR19 
    &111.66&120.80
    & .7407 & .8238 & .8023 & .8782
    & .9087 & .8897 & .8917 & .0132
    & .8068 & .9203 & .8747 & .0439
    \\
    ITSD-R
    \cite{itsd}
    & CVPR20 
    &26.07&15.96
    & .7227 & .8228 & .7973 & .8689
    & .9236 & .9040 & .8875 & .0116
    &.8148 & .9204 & .8723 & .0459
    \\
    GateNet-R
    \cite{gatenet}
    & ECCV20 
    &128.63&162.19
    & .6776 & .8307 & .7923 & .8338
    & .9283 & .8963 & .8890 & .0118
    & .8006 & .9212 & .8728 & .0448
    \\
    U2Net 
    \cite{u2net}
    & PR20 
    &44.01&58.83
    & .7154 & .8347 & .8045 & .8658
    & .9217 & .9013 & .8950 & .0133
    &.7942 & .9050 & .8610 & .0497
    \\
    PoolNet+ 
    \cite{poolnet+}
    & TPAMI21 
    &68.26&676.99
    & .7485 & .8565 & .8285 & .8858
    & .9337 & .9126 & .9071 & .0108
    & .8082 & .9203 &.8746 & .0415
    \\
    PFSNet 
    \cite{pfsnet}
    & AAAI21 
    &31.18&45.50
    & .8409 & .8616 & .8517 & .9343
    & .9341 & .9287 & .9071 & .0100
    & .8322 &.9215 & .8795 & .0405
    \\
    ICON-R 
    \cite{SOD_ICON}
    & TPAMI22 
    &33.04&20.97
    & .8242 & .8539 & .8402 & .9305
    & .9320 & .9260 & .9034 & .0104
    & .8280 & .9231 & .877 & .0420
    \\
    ELSANet
    \cite{ELSANet_TCSVT2023}
    & TCSVT23
    & 31.92 & 21.84
    & .8514 & .8607 & .8533 & .9381
    & .9371 & .9333 & .9032 & .0095
    & .8449 & .9257 & .8790 & .0387
    \\
    DCNet-R
    \cite{zhu2025dc}
    & PR25
    & 83.40 & 98.43
    & .8242 & .8658 & .8480 & .9318
    & .9370  & .9294 & \textcolor{green}{.9140} & .0104
    & .8413 & .9269 & .8828 & .0402
    \\
    M3Net-R 
    \cite{m3net}
    & -
    &34.61&55.49
    & .8041 & .8288 & .8166 & .9248
    & .9265 & .9198 & .8863 & .0115
    & .8404 & .9269 & .8833 & .0421
    \\ \hline
    \multicolumn{16}{c}{\textbf{Transformer-Based}}
    \\ \hline
    VST 
    \cite{SOD_VST}
    & ICCV21 
    &44.08&23.24
    & .7313 & .8143 & .7934 & .8807
    & .9153 & .8987 & .8832 & .0115
    &.8203 & .9314 & .8847 & .0418
    \\
    ICON-S 
    \cite{SOD_ICON}
    & TPAMI22
    &92.15&52.80
    & .8396 & .8623 & .8502 & .9379
    & .9387 & .9331 & .9071 & .0094
    & .8770 & \textcolor{green}{.9527} & .9049 & .0292
    \\
    CTIFNet
    \cite{yuan2023ctif}
    & TCSVT23
    & 329.17 & 65.84
    & .7601 & .7590 & .7532 & .8747 
    & .8749 & .8609 & .8360 & .0119
    & .8291 & .9497 & .8988 & .0352
    \\
    SelfReformer 
    \cite{SelfReformer}
    & TMM23 
    &91.58&18.61
    & .7521 & .8561 & .8274 & .8910
    & .9373 & .9229 & .9065 & .0098
    & .8633 & .9458 & .8988 & .0316
    \\
    BBRF 
    \cite{BBRF}
    & TIP23 
    &74.01&46.38
    & \textcolor{blue}{.8613} & .8670 & .8609 & .9376
    & .9376 & .9337 & .9086 & .0089
    &.8756 & .9441 & .8952 & .0305
    \\
    GLSTR
    \cite{ren2024unifying}
    & TETCI24
    & 162.54 & 372.50
    & .6532 & .8082 & .7642 & .8189
    & .9233 & .8831 & .8766 & .0129
    & .8635 & .9504 & .9072 & .0311
    \\
    GPONet
    \cite{yi2024gponet}
    & PR24
    & 71.46 & 145.39
    & .6638 & .8382 & .8017 & .8194
    & .9256 & .9006 & .8864 & .0138
    & .8619 & .9456 & .9053 & .0330
    \\
    TransformerSOD
    \cite{TransformerSOD_TCSVT2024}
    & TCSVT24
    & 87.11 & 47.72
    & .8097 & .8613 & .8411 & .9279 
    & \textcolor{green}{.9438} & .9387 & .9005 & .0093
    & .8786 & .9514 & .9059 & .0294
    \\
    DCNet-S
    \cite{zhu2025dc}
    & PR25
    & 509.61 & 211.27
    & .8173 & .8625 & .8447 & .9291
    & .9390 & .9314 & .9112 & .0096
    & .8834 & \textcolor{blue}{.9537} & \textcolor{green}{.9110} & \textcolor{blue}{.0281}
    \\
    M3Net-S 
    \cite{m3net}
    & -
    &102.75&381.60
    & .8515 & \textcolor{green}{.8732} & \textcolor{green}{.8622} & \textcolor{blue}{.9433}
    & \textcolor{blue}{.9463} & \textcolor{blue}{.9413} & .9122 & \textcolor{green}{.0090}
    & \textcolor{blue}{.8905} & .9516 
    & \textcolor{blue}{.9127} & .0287
    \\ \hline
    \multicolumn{16}{c}{\textbf{Mamba-Based}}
    \\ \hline
    VMamba-B \cite{liu2024vmamba}
    & NeurIPS24
    & 100.31 &60.82
    & \textcolor{green}{.8553} & \textcolor{blue}{.8751} & \textcolor{blue}{.8642} & \textcolor{green}{.9405}
    & .9422 & \textcolor{green}{.9389} & \textcolor{blue}{.9176} & \textcolor{blue}{.0085}
    & \textcolor{green}{.8895} & .9513 & .9103 & \textcolor{green}{.0284}
    \\
    \textbf{Tramba(Ours)} 
    &-
    & 111.44 & 72.38
    & \textcolor{red}{.8694} & \textcolor{red}{.8889}
    & \textcolor{red}{.8783} & \textcolor{red}{.9494}
    & \textcolor{red}{.9508} & \textcolor{red}{.9474}
    & \textcolor{red}{.9253} & \textcolor{red}{.0076}
    & \textcolor{red}{.8980} & \textcolor{red}{.9544} & \textcolor{red}{.9145} & \textcolor{red}{.0267}
    \\ \bottomrule[2pt]
\end{tabular}
\label{Quantitative}
\end{table*}

\begin{figure*}[!tb]
    \centering
    \includegraphics[width=.92\linewidth]{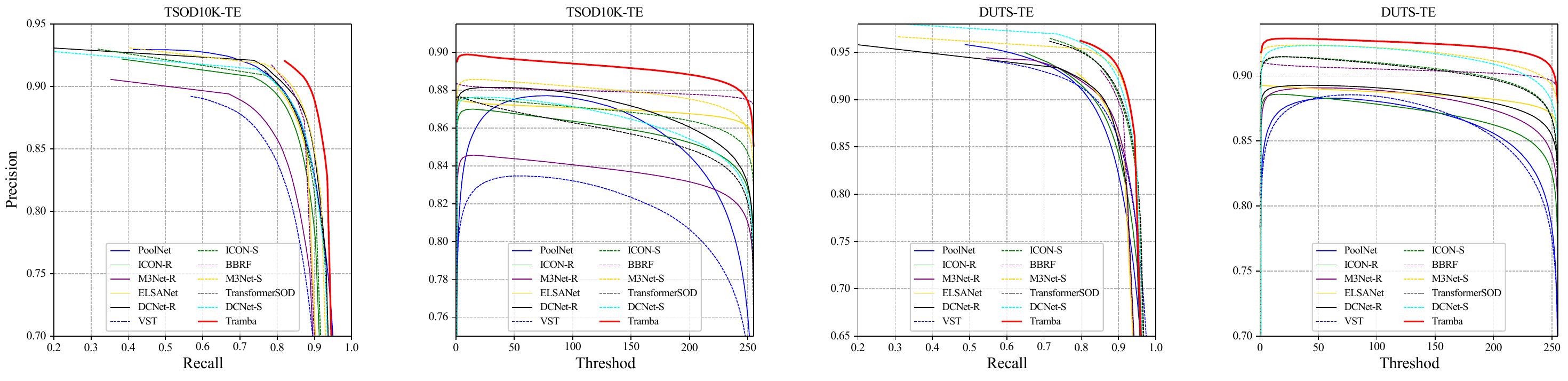}
    \caption{PR and F-measure curves of Tramba and SOTA NSI-SOD methods on TSOD10K-TE and DUTS-TE\cite{DUTS} datasets.}
    \label{fig:pr}
\end{figure*}

\begin{figure*}[tb]
    \centering
    \includegraphics[width=\linewidth]{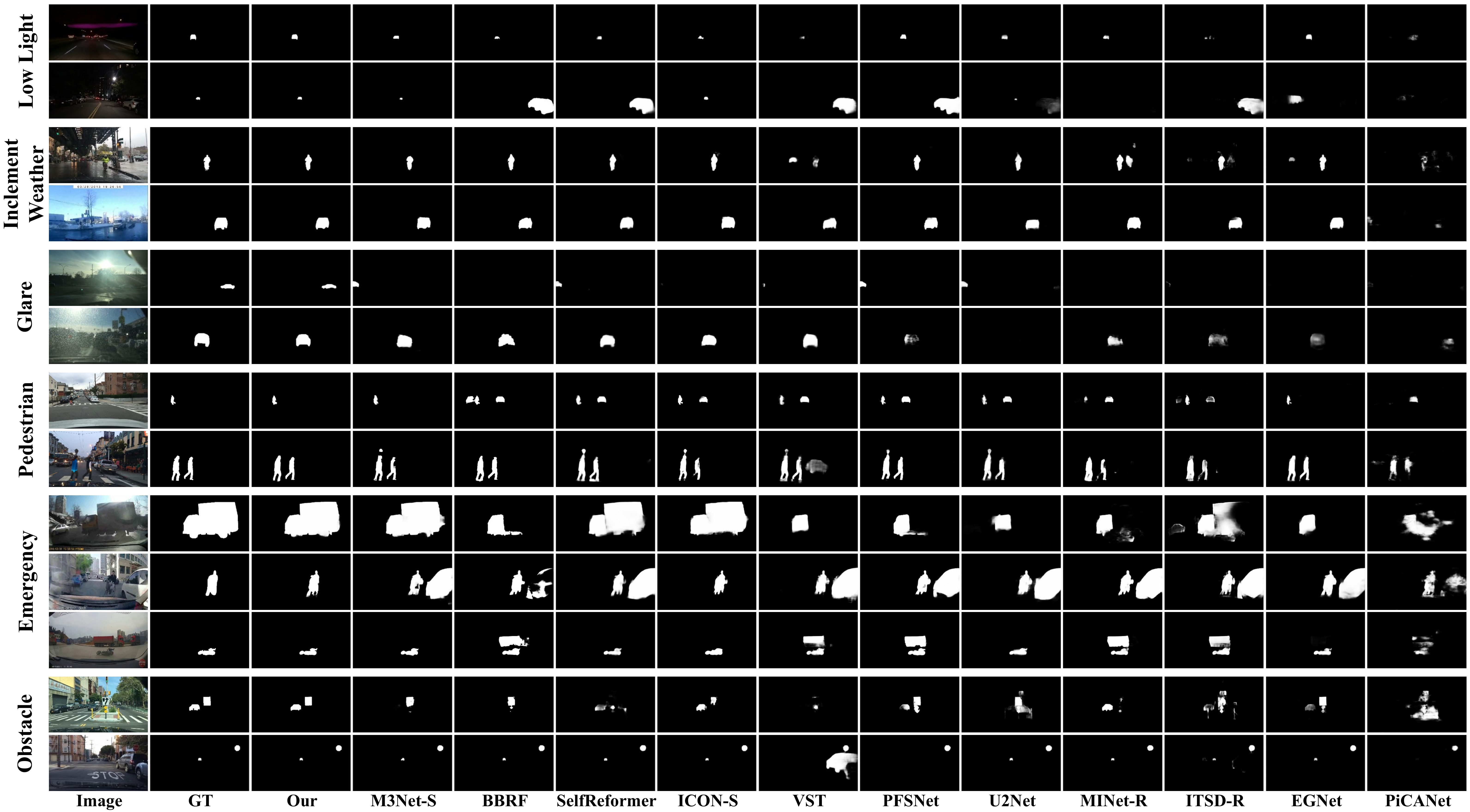}
    \caption{Qualitative comparison with 11 SOTA models in representative and challenging traffic targets and scenarios.}
    \label{fig:qualitative}
\end{figure*}

To effectively address the directional deficiency of traditional SS2D and enhance focus on key areas of traffic scenes, we propose a novel Helix Visual State Space (HVSS) module (Fig. \ref{fig:overal}). HVSS is inspired by the human visual system's tendency to prioritize the center of the visual field for enhanced balance \cite{gao2007discriminant}. Deng \etal \cite{deng2016does} also confirmed that experienced drivers primarily focus on the road ahead to ensure safe driving. Similarly, statistical analysis in Fig. \ref{fig:attributes} also shows that salient objects are concentrated in the central field of view.
Specifically, in Fig. \ref{fig:helix}, Helix SS2D uses the designed Helix Scan to involve scanning short sequences along a central axis, which can be regarded as an ``image slice'' in one direction, with each slice covering the central region. After each scan, the axis rotates by two patch distances, continuing until a full rotation is completed, forming two sequences of forward and backward slices. The remaining gaps provide another pair. Then, we use Selective SSM(S6) blocks to parallel capture the contextual information of each sequence and merge output sequences, ensuring comprehensive integration of contextual information from all directions at each position. Additionally, the Cross Scan\cite{vmamba} is incorporated as a supplement to the horizontal and vertical orientation information.
The process of Helix Scan can be formalized as:
\begin{equation}
\begin{split}
    &\overrightarrow{\mathbf{X}_{\mathrm{h}}} =  {\textstyle \bigcup_{k=0}^{\left \lfloor \frac{H}{2}  \right \rfloor -1}}\left \{ \mathbf{X}_{i,j} |(i,j) \in \overrightarrow{S_{k}^{1}} \cup  \overrightarrow{S_{k}^{2}} \right \},\\
    &\overrightarrow{{\mathbf{X}_{\mathrm{h}}}'} =  {\textstyle \bigcup_{k=0}^{\left \lfloor \frac{H}{2}  \right \rfloor -1}}\left \{ \mathbf{X}_{i,j} |(i,j) \in \overrightarrow{S_{k+\frac{1}{2}}^{1}} \cup  \overrightarrow{S_{k+\frac{1}{2}}^{2}} \right \},\\
    &\overleftarrow{\mathbf{X}_{\mathrm{h}}}=\left \{ \overrightarrow{\mathbf{X}_{\mathrm{h}}}_{(N-1-i)} \right \}_{i=0}^{N-1}, \overleftarrow{{\mathbf{X}_{\mathrm{h}}}'}=\left \{ \overrightarrow{{\mathbf{X}_{\mathrm{h}}}'}_{(N-1-i)} \right \}_{i=0}^{N-1},
    \label{e9}
\end{split}
\end{equation}
where $\overrightarrow{S_{k}^{1}}$ represents the set of points from $(2k,\!0)$ to $(H\!-\!1\!-\!2k,\!W-1)$, and $\overrightarrow{S_{k}^{2}}$ represents the set of points from $(H\!-\!1,\!2k)$ to $(0,\!W\!-\!1\!-\!2k)$ generated by Bresenham algorithm.

To improve multi-scale traffic target representation, HVSS introduces the depthwise multi-scale feedforward network (MS-FFN), which extracts multi-scale information using depthwise convolutions with 3×3, 5×5, and 7×7 kernels.

\begin{figure*}[tb]
    \centering
    \includegraphics[width=.95\linewidth]{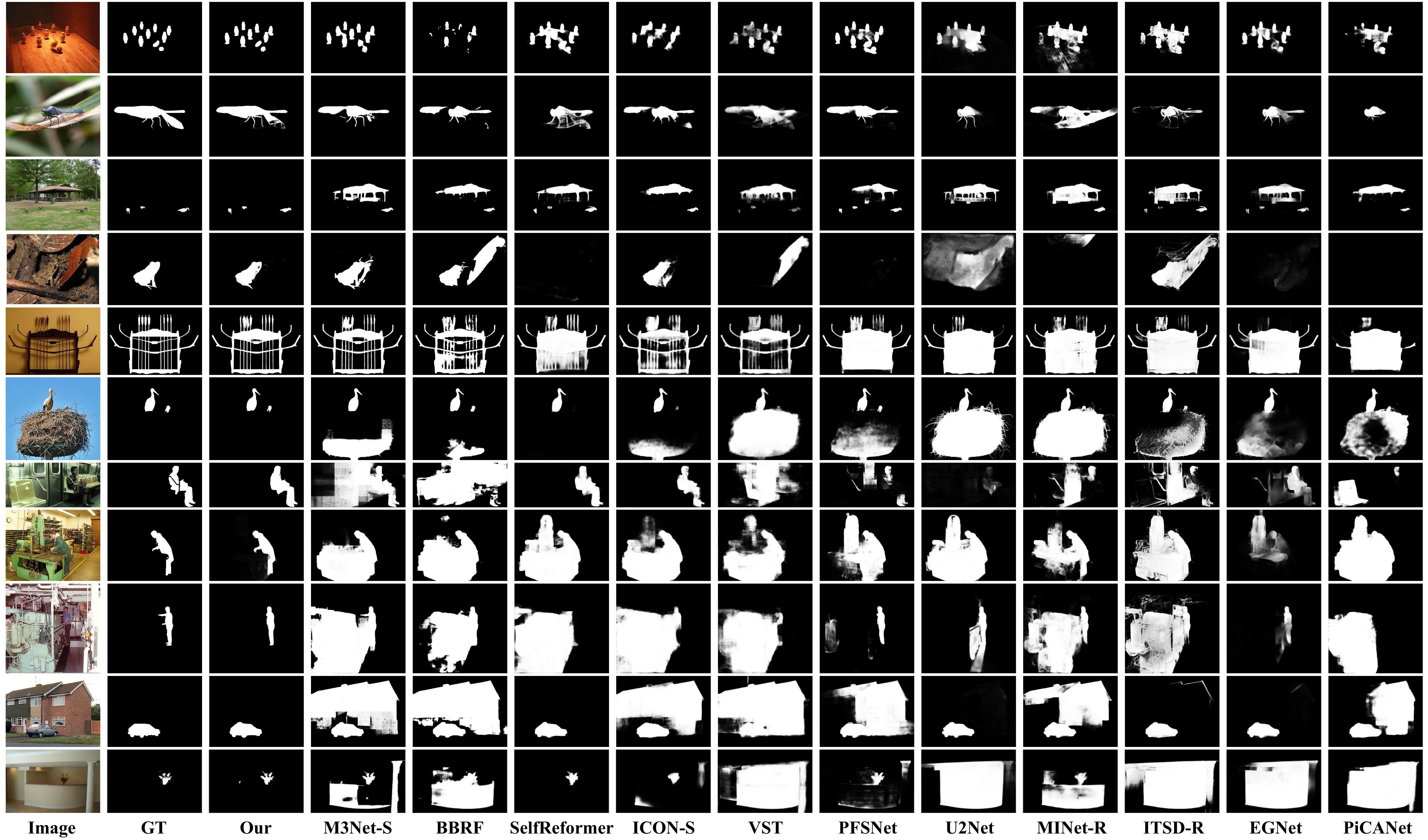}
    \caption{Additional qualitative comparison of our method with 11 NSI-SOD methods in some challenging scenarios in NSI-SOD.}
    \label{fig:sod_qualitative}
\end{figure*}

\section{Experiments}
\label{sec:experiments}

\subsection{Experimental Settings}
\textbf{Details.} We implement Tramba with PyTorch on NVIDIA RTX 4090 GPU (24 GB). During training, \revise{the encoder weights are initialized with pre-trained VMamba-B\cite{vmamba}.} Each image is resized to $384\times384$, and data augmentation methods (\eg cropping and flipping) are applied. We train the model using equally weighted binary cross-entropy (BCE) and intersection over union (IoU) loss, with deep supervision applied at each output stage. The Adam optimizer is employed with an initial learning rate of $1e^{-4}$, maintained for the first 60 epochs and subsequently reduced by a factor of 5 for the remaining 20 epochs, over a total of 80 training epochs.

\textbf{Criteria}. For TSOD evaluation, We employed four widely used metrics to quantitatively evaluate the performance of all the approaches, including F-measure score ($F_{\beta}$), Mean Absolute Error ($\mathcal{M}$), S-measure ($S_{\alpha } $), and E-measure ($E_{\xi } $). For both the F-measure and E-measure, we considered their maximum, mean, and adaptive forms (\textit{i.e.,} $F_{\beta}^{max}$, $F_{\beta}^{mean}$ and $F_{\beta}^{adp}$, $E_{\xi}^{max}$, $E_{\xi}^{mean}$ and $E_{\xi}^{adp}$). In detail, $F_{\beta}$ is defined as the weighted harmonic average of precision and recall, where $\beta^{2}$ is set to 0.3 to emphasize precision over recall. $\mathcal{M}$ is utilized to measure the pixel-wise average absolute difference between \revise{the} prediction and ground truth. $S_{\alpha}$ can simultaneously evaluate region-aware and object-aware structural similarities. $E_{\xi }$ jointly captures image-level statistics and local pixel-level matching information. For SOD evaluation, we follow previous studies and use weighted F-measure$F_{\beta}^{w}$, $E_{\xi}^{max}$, $S_{m}$, and $\mathcal{M}$.

\subsection{Comparison with State-of-the-arts}
Since there are currently no direct competitors for the TSOD task, we compare the proposed method with 23 SOTA NSI-SOD models, including 1) 12 CNN-based models including PiCANet \cite{picanet}, CPD \cite{cpd}, EGNet \cite{egnet}, ITSD \cite{itsd}, GateNet \cite{gatenet}, U2Net \cite{u2net}, PoolNet \cite{poolnet+}, PFSNet \cite{pfsnet}, ICON-R \cite{SOD_ICON}, ELSANet \cite{ELSANet_TCSVT2023}, DCNet-R \cite{zhu2025dc}, M3Net-R \cite{m3net}; 2) 10 Transformer-based models including VST \cite{SOD_VST}, ICON-S \cite{SOD_ICON}, CTIFNet \cite{yuan2023ctif}, SelfReformer \cite{SelfReformer}, BBRF \cite{BBRF}, GLSTR \cite{ren2024unifying}, GPONet \cite{yi2024gponet}, TransformerSOD \cite{TransformerSOD_TCSVT2024}, DCNet-S \cite{zhu2025dc} and M3Net \cite{m3net}; 3) a Mamba-based model VMamba \cite{liu2024vmamba}. 

To ensure fairness, all competing methods are fully trained on the \textbf{TSOD10K-TR} dataset until the loss is stable, based on their public source codes and default network parameter settings. In addition, to better prove the robustness and generalization ability of our Tramba, we retrain Tramba on the widely used NSI-SOD task training dataset \textbf{DUTS-TR}\cite{DUTS} and conduct a new comparison with these NSI-SOD models on \textbf{DUTS-TE}\cite{DUTS} dataset.

\textbf{Quantitative Comparison on TSOD and NSI-SOD.} To compare our Tramba with NSI-SOD state-of-the-art approaches, we conducted the quantitative evaluation on \textbf{TSOD10K-TE} (for TSOD comparison) and \textbf{DUTS-TE}\cite{DUTS} (for NSI-SOD comparison) datasets in terms of the evaluation metrics $F_{\beta}^{max}$, $F_{\beta}^{mean}$, $F_{\beta}^{adp}$, $E_{\xi}^{max}$, $E_{\xi}^{mean}$, $E_{\xi}^{adp}$, $S_{\alpha}$, and $\mathcal{M}$. The evaluation results are summarized in Tab. \ref{Quantitative}. 
The complexity comparison in terms of the number of parameters and FLOPs is also presented. Obviously, our method(Tramba) outperforms all CNN- or Transformer-based SOTA SOD models across all metrics for both tasks. Specifically, compared to the Mamba-based method VMamba\cite{vmamba}, Tramba demonstrates a clear performance advantage. 
Specifically, our Tramba outperforms CNN- and Transformer-based baselines by an average of at least 1\% on eight different metrics. Compared to the best-performing M3Net \cite{m3net} in NSI-SOD, our Tramba outperformed it by 2.10\%($F_{\beta}^{adp}$), 1.80\%($F_{\beta}^{max}$), 1.87\%($F_{\beta}^{mean}$), 0.65\%($E_{\xi}^{adp}$), 0.48\%($E_{\xi}^{max}$), 0.65\%($E_{\xi}^{mean}$), 1.44\%($S_{m}$), and 15.5\%($\mathcal{M}$) across eight different metrics on TSOD10K-TE dataset, respectively. This showcases its remarkable decoupling capability across two tasks that overlap yet exhibit significant differences.
In addition, we present the Precision-Recall (PR) and F-measure curves in Fig \ref{fig:pr}. The higher the curve is, the better the performance. As can be seen, the proposed Tramba favorably outperforms the other counterparts. 
These outstanding results demonstrate that Tramba not only effectively mimics human visual attention to salient regions but also distinguishes semantic priorities between saliency and danger effectively.

\textbf{Qualitative Comparison on TSOD. }
To more intuitively illustrate the advantages of the proposed algorithm, 
we visualize the results of Tramba and NSI-SOD methods in several challenging traffic scenarios in Fig. \ref{fig:qualitative}, 
including low light, inclement weather, glare, Pedestrian, Emergency, and Obstacle. 
Overall, benefiting from the DFVSS module's ability to capture global structures and local details, as well as the multi-perspective perception of HVSS, our Tramba can accurately locate the position of traffic salient objects, and preserve the completeness of objects with fine edge details, resulting in high-quality result maps in various traffic scenarios.

\textbf{Qualitative Comparison on NSI-SOD.} 
We also report the comparison on NSI-SOD. Fig. \ref{fig:sod_qualitative} presents the qualitative comparison results of Tramba-B with 11 other state-of-the-art NSI-SOD methods on challenging images from the DUTS-TE dataset. Our method achieves accurate localization and detailed segmentation of salient objects even in complex scenarios, such as multiple small targets (rows 1 and 3) and cluttered scenes (rows 8 and 9), demonstrating the robustness and generality of our Tramba in NSI-SOD field.

\subsection{Ablation Study}
\textbf{Effectiveness of Key Components.} We first evaluate the effect of our main designs. We start with the model without DFVSS and HVSS (``Baseline''), as shown in the first row of Tab. \ref{tab:eval_conponents}. Next, we sequentially add the proposed HFVSS, LFVSS, and Helix-SS2D, whose results are shown in the second $\sim$ fourth row of Tab. \ref{tab:eval_conponents}. We can see that 1) using the frequency features as the auxiliaries can bring an obvious performance improvement, 2) the proposed HFVSS (based on Window-SS2D) and LFVSS (based on Dilation-SS2D) effectively learn rich frequency-aware information that incorporates both high- and low-frequency cues, and 3) Helix-SS2D in HVSS enriches orientation encoding and central region emphasis, effectively enhancing model performance in complex traffic scenarios.

\begin{figure}[!tb]
    \centering
    \includegraphics[width=.9\linewidth]{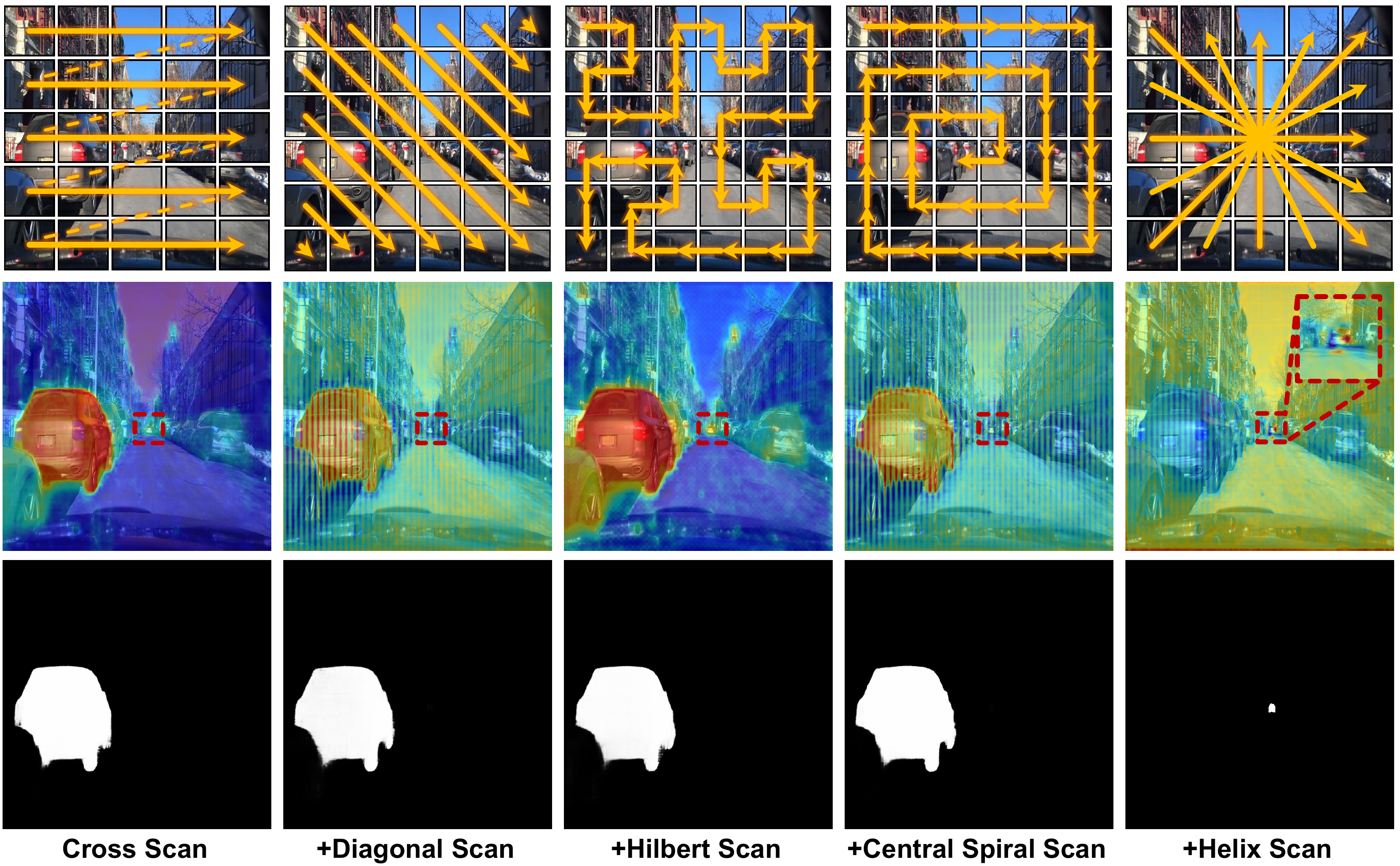}
    \caption{Heat maps of different scanning strategies.}
    \label{fig:hotmap}
\end{figure}

\textbf{Helix with other scanning strategies.} Tab. \ref{tab:eval_scan} reports the comparison results between the Helix-Scan and other scanning strategies within the HVSS module under individual environments, with the heatmaps visualized in \figref{fig:hotmap}. As expected, our Helix-Scan demonstrates the best performance. Parked cars pose minimal safety risks, whereas the driving status of visually inconspicuous vehicles ahead requires critical attention. Obviously, our Helix scan can exactly locate the vehicle ahead by filtering out visually prominent misleading elements, while others fail.

\setlength{\tabcolsep}{0.75mm}
\begin{table}[!tb]\small
\centering
\caption{Ablation for the effectiveness of key components.}
\begin{tabular}{c|cc|c||cccc} \toprule[1pt]
    \multirow{2}{*}{ID}&\multicolumn{2}{c|}{DFVSS} & \multicolumn{1}{c||}{HVSS} & \multicolumn{4}{c}{TSOD10K-TE}
    \\ \cline{2-8} 
    &\textit{HFVSS} & \textit{LFVSS}
    &\textit{Helix-SS2D}
    & $F_{\beta}^{adp}$
    & $E_{\xi}^{adp}$
    & $S_{m}$
    & $\mathcal{M}$
    \\ \hline
    1&& & 
    & .8553 & .9405 & .9176 & .0085
    \\
    2&\checkmark&   & 
    & .8596 & .9421 & .9194 & .0082
    \\
    3&\checkmark&\checkmark&
    & .8646 & .9453 & .9214 & .0078
    \\
    4&&&\checkmark
    &.8612 &.9435 & .9199 &.0080
    \\
    5&\checkmark& \checkmark  & \checkmark
    & \textbf{.8694} & \textbf{.9494} & \textbf{.9253} & \textbf{.0076}
    \\ \bottomrule[1pt]
\end{tabular}
\label{tab:eval_conponents}
\end{table}

\setlength{\tabcolsep}{1.25mm}
\begin{table}[!tb]\small
\centering
\caption{Ablation for the impact of different scanning.}
\begin{tabular}{c|c||cccc} \toprule[1pt]
    \multirow{2}{*}{ID}&\multirow{2}{*}{Scanning Method}  & \multicolumn{4}{c}{TSOD10K-TE}
    \\ \cline{3-6} 
    && $F_{\beta}^{adp}$
    & $E_{\xi}^{adp}$
    & $S_{m}$
    & $\mathcal{M}$
    \\ \hline
    1&Cross Scan\cite{vmamba}
    &.8553 &.9405 &.9176 &.0085
    \\
    2&+Diagonal Scan\cite{zhao2024rs}
    &.8588 &.9414 &.9183 &.0083
    \\
    3&+Hilbert Scan\cite{he2024mambaad}
    &.8568 &.9409 &.9177 &.0084
    \\
    4&+Central Spiral Scan\cite{wang2024soft}
    &.8576 &.9413 &.9172 &.0084
    \\ 
    5&\textbf{+Helix Scan(Ours)}
    &\textbf{.8612} &\textbf{.9435} &\textbf{.9199} &\textbf{.0080}
    \\ \bottomrule[1pt]
\end{tabular}
\label{tab:eval_scan}
\end{table}

\setlength{\tabcolsep}{1.75mm}
\begin{table}[!tb]\normalsize
\centering
\caption{Ablation for the impact of state space dimension.}
\begin{tabular}{c|c||cccc} \toprule[1pt]
    \multirow{2}{*}{Dimension} & \multicolumn{1}{c||}{FLOPs} & \multicolumn{4}{c}{TSOD10K-TE}
    \\ \cline{3-6} 
    & (G)
    & $F_{\beta}^{adp}$ & $E_{\xi}^{adp}$
    & $S_{m}$ & $\mathcal{M}$
    \\ \hline
    $d\_ state = 1$ & 71.45
    & .8696  & .9469 & .9244 & .0079
    \\
    $d\_ state = 8$ & 72.38
    & \textbf{.8694} & \textbf{.9494} & \textbf{.9253} & \textbf{.0076}
    \\ 
    $d\_ state = 16$  & 73.43
    & {.8702}  & .9478 & {.9253} & {.0076}
    \\ \bottomrule[1pt]
\end{tabular}
\label{tab:dimension}
\end{table}

\begin{figure*}[!tb]
\centering
\subfloat[]{\includegraphics[width=.8\linewidth]{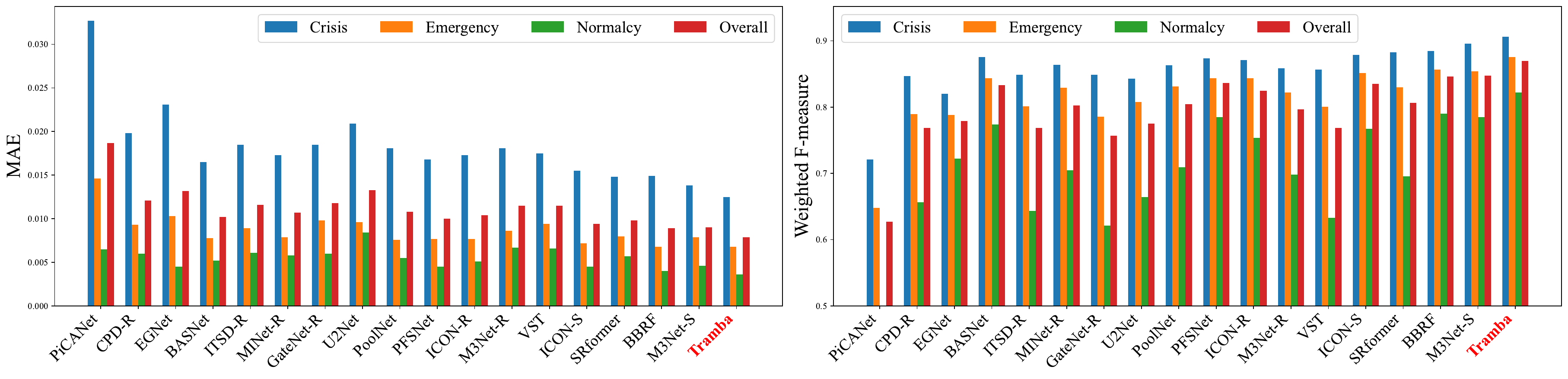}} \\
\subfloat[]{\includegraphics[width=.8\linewidth]{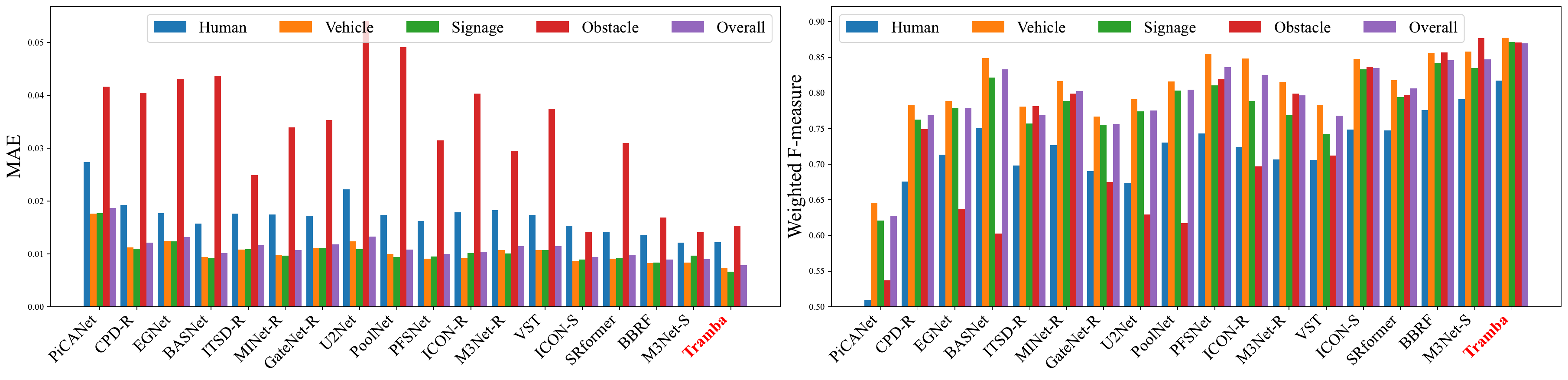}} \\
\subfloat[]{\includegraphics[width=.8\linewidth]{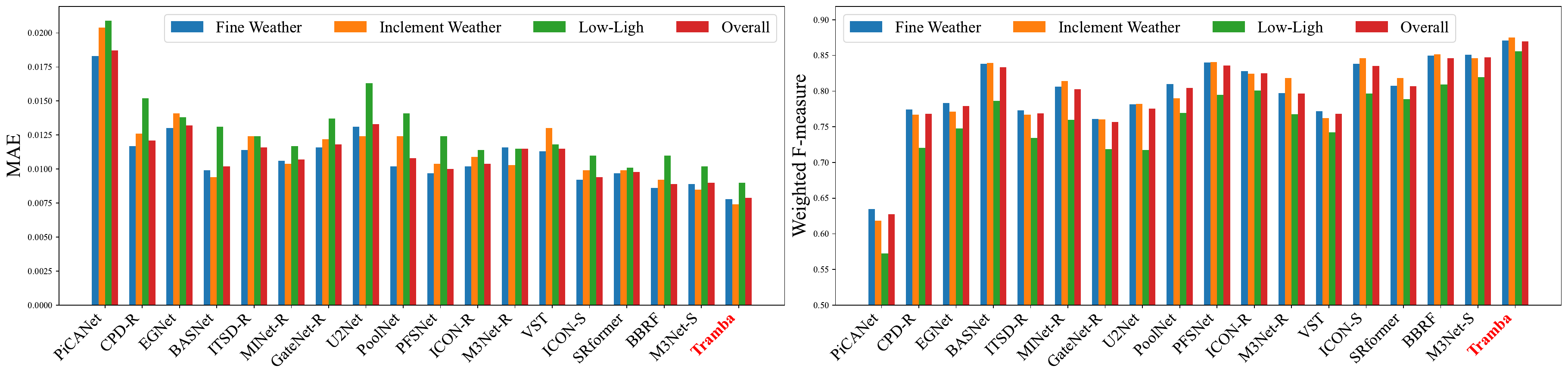}} \\
\subfloat[]{\includegraphics[width=.8\linewidth]{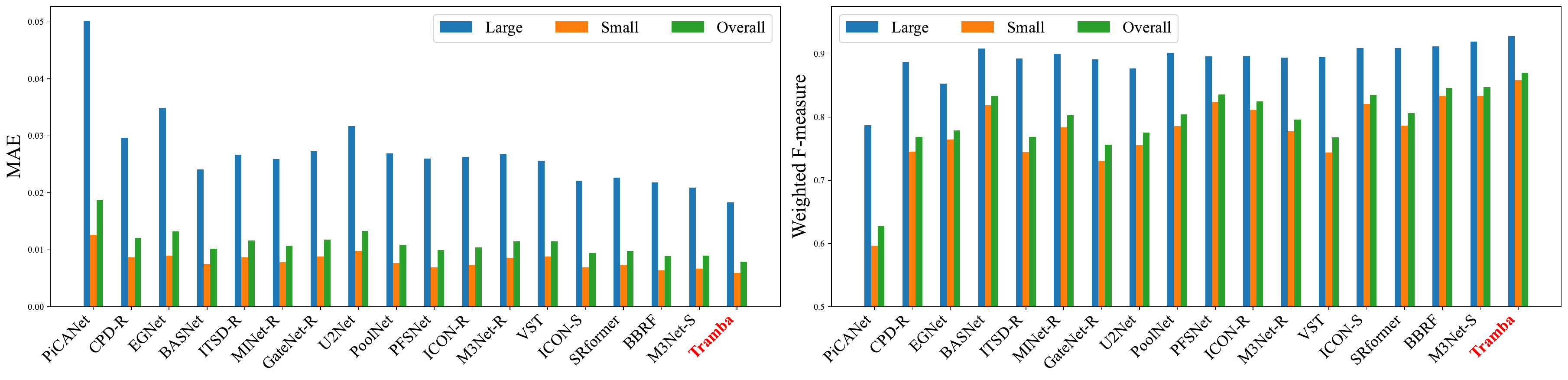}}
\caption{Evaluation on different class subsets of TSOD10K-TE. (a) Emergency levels: crisis, emergency, and normalcy. (b) Object categories: vehicle, human, signage, and obstacles. (c) Weather conditions: fine, inclement, and low-light. (d) Object sizes: large and small.}
\label{fig:ablation_subset}
\end{figure*}

\textbf{Impact of Hidden State Space Dimension.} 
We also conduct additional experiments to investigate the impact of state space's dimension ($d\_ state$) on model performance, as shown in Tab. \ref{tab:dimension}. With all other hyperparameters frozen, progressively increasing $d\_ state$ from 1 to 8 yields a slight performance improvement. Increasing it further to 16 shows that a larger dimension doesn't necessarily lead to a significant performance gain.

\textbf{Cross-Dataset Training-Testing.}
We conduct cross-dataset training-testing experiments to evaluate the differences between TSOD and NSI-SOD tasks. To ensure fairness, the datasets for both tasks are standardized in scale, with 10,000 training images and 2,500 testing images selected through random sampling. As shown in Tab. \ref{tab:dataset}, directly transferring a model trained on TSOD to NSI-SOD (or vice versa, in $1^{st}$/$2^{nd}$ row) results in significant performance gaps compared to in-situ training-testing, indicating substantial differences between the TSOD and NSI-SOD tasks. 

\setlength{\tabcolsep}{0.65mm}
\begin{table}[!tb]\small
\centering
\caption{Results for cross-dataset training-testing on randomly sampled TSOD10K-TR/TE (TSOD) and DUTS-TR/TE (NSI-SOD).}
\begin{tabular}{c||ccc|ccc} \toprule[1pt]
    \multirow{2}{*}{Dataset} & \multicolumn{3}{c|}{\textit{test on} TSOD} & \multicolumn{3}{c}{\textit{test on} NSI-SOD}
    \\ \cline{2-7} 
    & $F_{\beta}^{w}$ & $E_{\xi}^{max}$
    & $S_{m}$
    & $F_{\beta}^{w}$ & $E_{\xi}^{max}$
    & $S_{m}$
    \\ \hline
    \textit{train on} TSOD
    & .8709 & .9509 & .9255 
    & .7812 & .8854 & .8417 
    \\
    \textit{train on} NSI-SOD 
    & .5918 & .7942 & .7821 
    & .8952 & .9531 & .9128 
    \\ \hline
    \textbf{Gap}
    & \textbf{.2791} & \textbf{.1567} & \textbf{.1434}
    & \textbf{.1140} & \textbf{.0677} & \textbf{.0711} 
    \\ \bottomrule[1pt]
\end{tabular}
\label{tab:dataset}
\end{table}

\textbf{Subset Evaluation on TSOD10K-TE.} 
To better understand the model's performance across various scenarios and to identify its strengths and weaknesses in specific contexts, we divide the dataset into several subsets based on their various attributes (\ie emergency levels, object categories, weather conditions, and object sizes). Then, we re-evaluate our Tramba with other SOTA NSI-SOD methods on metrics MAE and weighted F-measure for these subsets, as shown in Fig. \ref{fig:ablation_subset}. As expected, objects classified as \textit{crisis}, \textit{human/obstacle}, \textit{low-light}, and \textit{small} are more challenging to detect. Nevertheless, our method achieves the best performance in the majority of the fine-grained categories.

\subsection{Failure Case Analysis}
\begin{figure}[!tb]
    \centering
    \includegraphics[width=.95\linewidth]{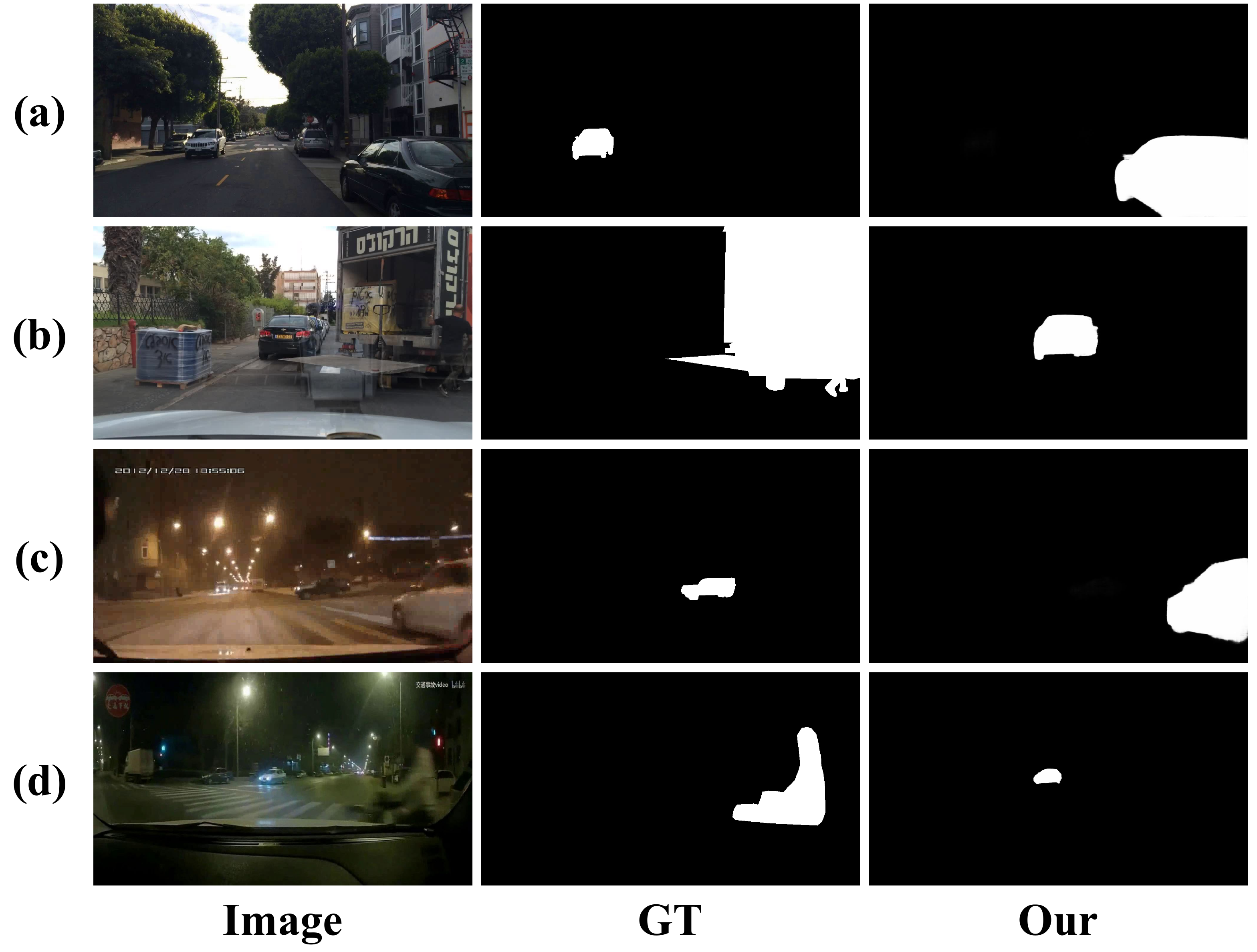}
    \caption{Some failure cases of the proposed Tramba.}
    \label{fig:failure}
\end{figure}

Although the proposed Tramba method performs exceptionally well on the TSOD task, rarely producing completely incorrect predictions, there are still some failure cases, as shown in Fig. \ref{fig:failure}. In the first row (a), our method is attracted to the parked vehicle by the roadside, overlooking the oncoming vehicle. In the second row (b), the truck is the more noteworthy target, rather than the black sedan. The third row (c) presents a nighttime traffic intersection scenario where vehicles traveling on the crossing lane require special attention, yet our method failed to detect it. Meanwhile, in the fourth row (d), our method fails to detect the motion-blurred object under the low-light condition. One potential reason for these failures is that in traffic scenarios, the model is easy to be misled by closer objects with stronger color contrast, leading it to overlook the targets that may have a significant impact on safety.

\section{Conclusion}
\label{sec:conclusion}
This paper presents a comprehensive study on traffic salient object detection (\textbf{TSOD}) and establishes safety-semantic saliency as a new paradigm for traffic scene understanding, addressing the critical limitations of visual-driven approaches. We provide the first challenging and densely annotated \textbf{TSOD10K} benchmark, covering a wide range of real-world traffic scenarios, including diverse road environments under various weather and lighting conditions.
We propose a Mamba-based TSOD model, named Tramba. Specifically, a novel Dual-Frequency Visual State Space (DFVSS) module equipped with Window-SS2D and Dilated-SS2D mechanisms is explored to enhance the perception of local details and global structures by incorporating high/low-frequency cues. Then, an innovative Helix-SS2D mechanism is proposed to integrate driving attention priors while capturing global multi-directional information. Extensive experiments demonstrate that Tramba achieves SOTA performance on both TSOD and NSI-SOD tasks. We hope this research can offer the community an opportunity to design new models for the intelligent transportation field. Future directions include extending the framework to multi-modal sensor fusion and real-time deployment in autonomous driving systems.

\section*{Acknowledgments}
This work was supported by the National Natural Science Foundation of China under Grant 62303173 and 62403256.

{
\small
\bibliographystyle{IEEEtran}
\bibliography{TrafficSOD}
}

\end{document}